\documentclass[
]{ceurart}

\usepackage{listings}
\input{commands.sty}

\newcommand{\oruga}{\textsc{Oruga}}
\newcommand{\crc}{%
	\begin{tikzpicture}[inner sep = 0pt,baseline]
	\draw (0,0) circle (.07cm);
	\fill[black, opacity = 0.15] (0,0) circle (.07cm);
	\end{tikzpicture}%
}
\newcommand{\oR}{%
	\begin{tikzpicture}[inner sep = 0pt,baseline]
	\node[anchor = east] (o1) at (0,0) {\crc};
	\node[anchor = east] (o2) at (6pt,0) {\crc};
	\node[anchor = east] (o1) at (12pt,0) {\crc};
	\node[anchor = east] (o1) at (18pt,0) {\crc};
	\node[anchor = east] (o1) at (0,6pt) {\crc};
	\node[anchor = east] (o2) at (6pt,6pt) {\crc};
	\node[anchor = east] (o1) at (12pt,6pt) {\crc};
	\node[anchor = east] (o1) at (18pt,6pt) {\crc};
	\node[anchor = east] (o1) at (0,12pt) {\crc};
	\node[anchor = east] (o2) at (6pt,12pt) {\crc};
	\node[anchor = east] (o1) at (12pt,12pt) {\crc};
	\node[anchor = east] (o1) at (18pt,12pt) {\crc};
	\end{tikzpicture}%
}
\newcommand{\oTsmall}{%
\begin{tikzpicture}[inner sep = 0pt,baseline]
\node[anchor = east] (o1) at (0,0) {\crc};
\node[anchor = east] (o2) at (6pt,0) {\crc};
\node[anchor = east] (o1) at (0,6pt) {\crc};
\end{tikzpicture}%
}
\newcommand{\oTwo}{%
\begin{tikzpicture}[inner sep = 0pt,baseline]
\node[anchor = east] (o1) at (0,0) {\crc};
\node[anchor = east] (o2) at (6pt,0) {\crc};
\end{tikzpicture}%
}
\newcommand{\oThree}{%
\begin{tikzpicture}[inner sep = 0pt,baseline]
\node[anchor = east] (o1) at (0,0) {\crc};
\node[anchor = east] (o2) at (6pt,0) {\crc};
\node[anchor = east] (o2) at (12pt,0) {\crc};
\end{tikzpicture}%
}
\newcommand{\oT}{%
	\begin{tikzpicture}[inner sep = 0pt,baseline]
	\node[anchor = east] (o1) at (0,0) {\crc};
	\node[anchor = east] (o2) at (6pt,0) {\crc};
	\node[anchor = east] (o1) at (12pt,0) {\crc};
	\node[anchor = east] (o1) at (0,6pt) {\crc};
	\node[anchor = east] (o2) at (6pt,6pt) {\crc};
	\node[anchor = east] (o1) at (0,12pt) {\crc};
	\end{tikzpicture}%
}
\newcommand{\oTrot}{%
	\rotatebox{180}{%
		\begin{tikzpicture}[inner sep = 0pt,baseline=12pt]
		\node[anchor = east] (o1) at (0,0) {\crc};
		\node[anchor = east] (o2) at (6pt,0) {\crc};
		\node[anchor = east] (o1) at (12pt,0) {\crc};
		\node[anchor = east] (o1) at (0,6pt) {\crc};
		\node[anchor = east] (o2) at (6pt,6pt) {\crc};
		\node[anchor = east] (o1) at (0,12pt) {\crc};
		\end{tikzpicture}%
	}%
}
\newcommand{\stt}[1]{{\scalebox{0.85}[0.9667]{\texttt{#1}}}}

\begin{document}
\copyrightyear{2022}
\copyrightclause{Copyright for this paper by its authors.
	Use permitted under Creative Commons License Attribution 4.0
	International (CC BY 4.0).}
	
\conference{HLC 2022: 3rd International Workshop on Human-Like Computing,
	September 28--30, 2022, Cumberland Lodge, UK}
	
\title{Oruga: An Avatar of Representational Systems Theory}


\author[1]{Daniel Raggi}[%
orcid=0000-0002-9207-6621,
email=daniel.raggi@cl.cam.ac.uk
]
\cormark[1]

\author[1]{Gem Stapleton}[%
orcid=0000-0002-6567-6752,
]

\author[1]{Mateja Jamnik}[%
orcid=0000-0003-2772-2532,
]

\author[2]{Aaron Stockdill}[%
orcid=0000-0003-3312-5267,
]

\author[2]{Grecia Garcia\ Garcia}[%
orcid=0000-0002-7327-7225,
]

\author[2]{Peter C.-H. Cheng}[%
orcid=0000-0002-0355-5955,
]

\address[1]{University of Cambridge, Cambridge, UK}
\address[2]{University of Sussex, Brighton, UK}
\cortext[1]{Corresponding author.}

	\begin{abstract}
		Humans use representations flexibly. We draw diagrams, change representations and exploit creative analogies across different domains. We want to harness this kind of power and endow machines with it to make them more compatible with human use.
		Previously we developed Representational Systems Theory (RST) to study the structure and transformations of representations~\cite{raggi2022rst,raggi2022transfer}. In this paper we present {\oruga} (\textit{caterpillar} in Spanish; a symbol of transformation), an implementation of various aspects of RST. {\oruga} consists of a core of data structures corresponding to concepts in RST, a language for communicating with the core, and an engine for producing transformations using a method we call \textit{structure transfer}. In this paper we present an overview of the core and language of {\oruga}, with a brief example of the kind of transformation that structure transfer can execute.
	\end{abstract}
\maketitle

\vspace{-0.1cm}
\section{Introduction: rep2rep and RST}
This work is part of the rep2rep project~\citep{jamnik:emwtehatsrwah,cheng2021cognitive}, whose aim is to study and implement systems that mimic and accommodate the flexibility of human representational skills.  As part of rep2rep, we developed Representational Systems Theory (RST)~\citep{raggi2022rst,raggi2022transfer}, a theoretical foundation for studying the structure and transformations of representations. The main requirement for such a theory is that it must be general enough to account for the diversity of representations used by humans (e.g., formal and natural languages, geometric figures, graphs, plots, etc.), and at the same time it must be rigorous and precise enough to be implementable. For example, it must be able to explain the relation between the validity of $1+2=3$ and the fact that {\,\oTsmall\,} can be built by joining \adjustbox{raise=3pt}{\,\oTwo\,} and \adjustbox{raise=3pt}{\,\crc}, and ultimately allow us to produce such transformations between arithmetic terms and dot diagrams; but its scope must not be limited to arithmetic and dot diagrams.

One of the key innovations of RST is the notion of a \textit{construction space}, where many concepts of interest for the study of representations can be defined in graph-theoretic terms. Notably, the concept of a \textit{construction} generalises that of a syntax tree, but its weaker restrictions and our graph-theoretic approach allow us to model more complex structures and inspect their properties. Importantly, it allows us to model representations often considered informal, and to do so uniformly across different representational systems so that we can encode relations and produce transformations between them.

In this paper we present an overview of {\oruga}'s core data structures and language for communicating with the core. Specifically, we demonstrate how some of the main concepts of RST are declared in {\oruga}. Here we do not focus on the engine for producing transformations. The Standard ML code can be found in~\cite{raggirep2rep}.


\section{The core of {\oruga}} 
{\oruga}'s core data structures are \textit{type systems}, \textit{constructor specifications}, \textit{constructions} and \textit{transfer schemas}. These are crucial for specifying construction spaces, building structures within them, and producing transformations across them.
\subsection{Type Systems}
In RST we refer to concrete representations as \textit{tokens}, and we assign them \textit{types}. This induces equivalence classes of tokens apropos to the token-type dichotomy~\cite{sep-types-tokens}. RST is agnostic concerning the criteria for determining whether two tokens have the same type -- it simply regards \textit{type} as a function that assigns a value to every token. For example, we may say that the arithmetic expression $1+1$ contains two tokens of type \texttt{one}. 
RST also enables subtyping via a partial order on the set of types. For instance, we can set the order to be such that \stt{one} is a subtype of \stt{numeral}, and \stt{numeral} is a subtype of \stt{numExp}. 
Formally, we define a \textit{type system} as a pair, $(\types,\leq)$, where $\types$ is a set whose elements are called types, and $\leq$ is a partial order over $\types$. 
\begin{wrapfigure}[6]{r}{6.7cm}
	\vspace*{-0.175cm}
	\adjustbox{minipage=7.2cm,fbox,scale={0.9}{1}}{%
		\begin{alltt}\scriptsize
			typeSystem arithT =
			\ \ types _:numeral, _:var, _:numExp, _:formula,
			\ \ \ \ \ \ \ \ plus, minus, binOp, leq, equals, binRel
			\ \ order var < numExp, numeral < numExp,
			\ \ \ \ \ \ \ \ plus < binOp, minus < binOp,
			\ \ \ \ \ \ \ \ leq < binRel, equals < binRel
	\end{alltt}}
\end{wrapfigure}
In the {\oruga} language, we can declare type systems, as demonstrated here (right): a type system for a fragment of arithmetic. 
Expressions such as \stt{\_:var} declare that the type \stt{var} has infinitely many subtypes which are not explicitly declared. In practice, it means that the user of {\oruga} can write \stt{t:A:var}, and this means \stt{t} is a token of type \stt{A}, which is a subtype of \stt{var}. The transitive/reflexive closure of the subtype relation is calculated in the background to facilitate minimal declarations.

\subsection{Constructor specifications}
A construction space is where we encode how tokens are constructed, for example, how $1+2$ relates to $1$, $+$ and $2$. Formally, a construction space is a triple $(\tsystemn,\cspecificationn,G)$ where $\tsystemn$ is a type system, $\cspecificationn$ is a \textit{constructor specification}, and $G$ is a \textit{structure graph}. We explain these below.

Formally, a constructor specification is a pair $(\constructors,\sig)$ where $\constructors$ is a set of elements called \textit{constructors} and $\sig$ is a function with domain $\constructors$ that, given a constructor, returns a pair $([\tau_1,\ldots,\tau_n],\tau)$, where $[\tau_1,\ldots,\tau_n]$ is a finite sequence of \textit{input} types and $\tau$ is an \textit{output} type. 
For example, a constructor that infixes a binary operator, \stt{infixOp}, may be defined so that $\sig(\stt{infixOp}) = ([\stt{numExp}, \stt{binOp}, \stt{numExp}],\stt{numExp})$. 

\begin{wrapfigure}[3]{r}{6.7cm}
	\vspace*{-0.85cm}
	\adjustbox{minipage=7.2cm,fbox,scale={0.9}{1}}{%
		\begin{alltt}\scriptsize
			conSpec arith:arithT =
			\ \ infixOp : [numExp,binOp,numExp] -> numExp,
			\ \ infixRel : [numExp,binRel,numExp] -> formula,
			\ \ implicitMult : [numExp,numExp] -> numExp
	\end{alltt}}
\end{wrapfigure}
See (right) how we declare a finite constructor specification \stt{arith} for type system \stt{arithT} in the {\oruga} language.

%
The structure graph associated with a construction space is the home of \textit{all} admissible constructions of \textit{every} token of the construction space. This is where the structure of representations is encoded. See~\cite{raggi2022rst} for a full formal definition. For most interesting construction spaces, its graph is infinite and perhaps undecidable (i.e., we cannot know if any arbitrary graph is a part of it). Then, insofar as it concerns implementation we need to think about manageable parts of structure graphs. This leads us to the concept of a \textit{construction}.
\subsection{Constructions and patterns}	
\begin{wrapfigure}[4]{r}{5.75cm}
	\vspace*{-1.3cm}
	\begin{tikzpicture}[construction]\scriptsize
	\node[termrep] (t) {$1+2=x$};
	\node[constructorNE = {\stt{infixRel}}, below = 0.35cm of t] (c) {};
	\node[termrep, below left = -0.2cm and 0.5cm of c] (t1) {$1+2$};
	\node[constructor = {\stt{infixOp}}, below = 0.35cm of t1,xshift=0.2cm] (c1) {};
	\node[termrep, below left = 0.3cm and 0.5cm of c1] (t11) {$1$};
	\node[termrep, below = 0.45cm of c1] (t12) {$+$};
	\node[termrep, below right = 0.3cm and 0.5cm of c1] (t13) {$2$};
	\node[termrep, below = 0.5cm of c] (t2) {\vphantom{j}$=$};
	\node[termrep, below right = 0.3cm and 0.3cm of c] (t3) {$x$};
	\path[->] (c) edge (t) 
	(t1) edge[bend left=10] node[index label]{1} (c) 
	(t2) edge node[index label]{2} (c) 
	(t3) edge[bend right=10] node[index label]{3} (c) 
	(c1) edge ([xshift=0.2cm]t1.south) 
	(t11) edge[bend left=10] node[index label]{1} (c1) 
	(t12) edge node[index label]{2} (c1)
	(t13) edge[bend right=10] node[index label]{3} (c1);
	\end{tikzpicture}\hspace{-0.5cm}
	\begin{tikzpicture}[construction]\scriptsize
	\node[termrep] (v) {{\scalebox{0.8}{\oT}}};
	\node[constructorpos = {\stt{rotate}}{130}{0.15cm}, left = 0.4cm of v] (u3) {};
	\node[termrep, below = 0.5cm of u3, xshift = 0.0cm, yshift = 0.0cm] (v9) {{\scalebox{0.8}{\oTrot}}};
	\node[constructorS = {\stt{remove}}, below = 0.35cm of v9,xshift=0.2cm] (u2) {};
	\node[termrep, left = 0.6cm of u2] (v8) {{\scalebox{0.8}{\oR}}};
	\path[->]
	(v8) edge[out =0, in = -180] node[index label] {1} (u2)
	(v9) edge[out = 90, in = -90] node[index label] {1} (u3)
	(v) edge[in=-0,out=-90,looseness=1.3] node[index label] {2} (u2)
	(u3) edge (v)
	(u2) edge ([xshift=0.2cm]v9.south);
	\end{tikzpicture}
\end{wrapfigure}
A construction captures \textit{one} way in which \textit{one} token is constructed.
	Here (right) we show two constructions, one for {$1+2=x$} and another for dot diagram\, \scalebox{0.75}{\oT}.
	
Constructions have many useful properties; in parti-

	\begin{wrapfigure}[7]{r}{6.44cm}
		\adjustbox{minipage=6.9cm,fbox,scale={0.9}{1}}{%
			\begin{alltt}\scriptsize
				construction con:arith =
				\ \ t:1plus2equalsx:formula
				\ \ \ \ <- infixRel[t1:1plus2:numExp
				\ \ \ \ \ \ \ \ \ \ \ \ \ \ \ \ \ \ \ <- infixOp[t11:1:numeral,
				\ \ \ \ \ \ \ \ \ \ \ \ \ \ \ \ \ \ \ \ \ \ \ \ \ \ \ \ \ \ t12:plus,
				\ \ \ \ \ \ \ \ \ \ \ \ \ \ \ \ \ \ \ \ \ \ \ \ \ \ \ \ \ \ t13:2:numeral],
				\ \ \ \ \ \ \ \ \ \ \ \ \ \ \ \ t2:equals, 
				\ \ \ \ \ \ \ \ \ \ \ \ \ \ \ \ t3:x:var]
		\end{alltt}}
	\end{wrapfigure}
%

\noindent
cular, they can be easily encoded with a recursive datatype (for implementation, see~\cite{raggirep2rep}).
See (right) how a construction is declared.
Notation \stt{t1:1plus2:numExp} creates a new type, \stt{1plus2}, such that \stt{1plus2} is a subtype of \stt{numExp}; this is allowed because we declared \stt{\_:numExp} in the type system.

A \textit{pattern} for a construction space, $(\tsystemn,\cspecificationn,G)$, is a construction that satisfies the restrictions

	\begin{wrapfigure}[7]{r}{3.4cm}
		\vspace{-0.4cm}
		\adjustbox{minipage=3.4cm}{%
	\begin{tikzpicture}[construction]\scriptsize
	\node[typeW = \stt{formula}] (t) {};
	\node[constructorNE = {\stt{infixRel}}, below = 0.35cm of t] (c) {};
	\node[typeNW={\stt{numExp}}, below left = 0.1cm and 0.4cm of c] (t1) {};
	\node[constructor = {\stt{infixOp}}, below = 0.35cm of t1] (c1) {};
	\node[typeS=\stt{one}, below left = 0.4cm and 0.5cm of c1] (t11) {};
	\node[typeS=\stt{binOp}, below = 0.45cm of c1] (t12) {};
	\node[typeS=\stt{numExp}, below right = 0.4cm and 0.6cm of c1] (t13) {};
	\node[typeSE=\stt{binRel}, below = 0.5cm of c] (t2) {};
	\node[typepos={\stt{numExp}}{35}{0.27cm}, below right = 0.3cm and 0.4cm of c] (t3) {};
	\path[->] (c) edge (t) 
	(t1) edge[bend left=10] node[index label]{1} (c) 
	(t2) edge node[index label]{2} (c) 
	(t3) edge[bend right=10] node[index label]{3} (c) 
	(c1) edge (t1) 
	(t11) edge[bend left=10] node[index label]{1} (c1) 
	(t12) edge node[index label]{2} (c1)
	(t13) edge[bend right=10] node[index label]{3} (c1);
	\end{tikzpicture}}
\end{wrapfigure}
\noindent
given by the type system and constructor specification, 
but may not necessarily be a part of $G$. Patterns are useful given the concept of \textit{matching}, as they allow us to capture classes of constructions. Roughly, we say that a construction matches a pattern if there exists an isomorphism from the former to the latter that respects the subtype order. See (right) an example of a pattern; the labels on token vertices specify types. The construction of $1+2=x$ (above) matches this pattern.
	\subsection{Transfer schemas}
	\begin{wrapfigure}[7]{r}{6.65cm}
		\vspace{-0.7cm}
			\begin{tikzpicture}[construction]\scriptsize
			\node[typeE = \stt{dotDiag}](t){t'};
			\node[constructorNE = \stt{join}, below = 0.35cm of t](u){};
			\node[typepos = {\stt{dotDiag}}{150}{0.45cm}, below left = 0.5cm and 0.2cm of u](t1){a};
			\node[typepos = {\stt{dotDiag}}{30}{0.4cm}, below right = 0.6cm and 0.2cm of u](t3){b};
			\path[->] (u) edge (t)
			(t1) edge[bend left = 10] node[index label]{1} (u)
			(t3) edge[bend right = 10] node[index label]{2} (u);
			\node[typeW = \stt{numExp}, left = 3.4cm of t](t'){t};
			\node[constructorNW= \stt{infixOp}, below = 0.35cm of t'](u'){};
			\node[typepos = {\stt{numExp}}{145}{0.35cm}, below left = 0.4cm and 0.2cm of u'](t1'){n};
			\node[typeS = {\stt{plus}}, below right = 0.45cm and -0.2cm of u'](t2'){p};
			\node[typepos = {\stt{numExp}}{35}{0.25cm}, below right = 0.45cm and 0.4cm of u'](t3'){m};
			\path[->] (u') edge (t')
			(t1') edge[bend left = 10] node[index label]{1} (u')
			(t2') edge[bend right = 5] node[index label]{2} (u')
			(t3') edge[bend right = 10] node[index label]{3} (u');
			\node[constructorS = $\stt{rep}$, below left = 0.2cm and 1.5cm of t] (uu) {};
			\node[typeIE = {\stt{true}}, above = 0.25cm of uu] (tt) {};
			\path[->,darkred] (uu) edge (tt)
			(t) edge[out = 180, in = 0] node[index label]{2} (uu)
			(t') edge[out = 0, in = 180] node[index label, pos = 0.5]{1} (uu);
			\node[constructorN = $\stt{rep}$, below left = 1.5cm and 1.3cm of uu] (uu1) {};
			\node[typeIE = {\stt{true}}, below = 0.2cm of uu1] (tt1) {};
			\path[->,darkblue] (uu1) edge (tt1)
			(t1) edge[out = -175, in = 5, looseness = 1.2] node[index label]{2} (uu1)
			(t1') edge[out = -100, in = 175, looseness = 1.3] node[index label]{1} (uu1);
			\node[constructorN = $\stt{rep}$, below right = 1.5cm and -0.2cm of uu] (uu2) {};
			\node[typeIE = {\stt{true}}, below = 0.2cm of uu2] (tt2) {};
			\path[->,darkblue] (uu2) edge (tt2)
			(t3) edge[out = 180, in = 10, looseness = 1.2] node[index label,pos=0.45]{2} (uu2)
			(t3') edge[out = -90, in = 175, looseness = 1.1] node[index label]{1} (uu2);
			\node[constructorN = $\stt{disj}$, below right = 1.5cm and 1.6cm of uu] (uu3) {};
			\node[typeIE = {\stt{true}}, below = 0.2cm of uu3] (tt3) {};
			\path[->,darkblue] (uu3) edge (tt3)
			(t1) edge[out = -120, in = 180, looseness = 1.2] node[index label,pos=0.45]{1} (uu3)
			(t3) edge[out = -40, in = 0, looseness = 1.4] node[index label]{2} (uu3);
			\end{tikzpicture}
	\end{wrapfigure}
One key concept for achieving transformations is that of a \textit{transfer schema}. A transfer schema is, roughly, an inference rule for deriving relations that cross construction spaces. We omit the formal definition here. See (right) a transfer schema which captures the fact that, provided that two
%
disjoint dot diagrams (\texttt{a} and \texttt{b}) represent two numerical expressions (\texttt{n} and \texttt{m}), then the result of joining them yields a representation of $\texttt{n}+ \texttt{m}$.

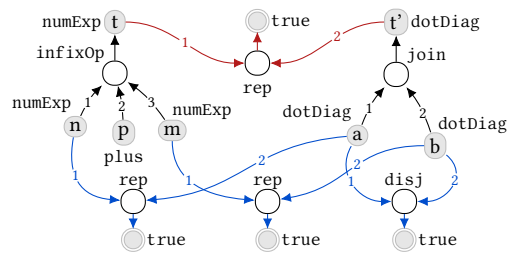
	\begin{wrapfigure}[6]{r}{5.9cm}
	\vspace{-0.0cm}
	\adjustbox{minipage=6.3cm,fbox,scale={0.9}{1}}{%
		\begin{alltt}\scriptsize
			tSchema plusJoin:(arith,dotDiagrams) =
			\ \ source t:numExp <- infixOp[n:numExp,
			\ \ \ \ \ \ \ \ \ \ \ \ \ \ \ \ \ \ \ \ \ \ \ \ \ \ \ \ \ p:plus,
			\ \ \ \ \ \ \ \ \ \ \ \ \ \ \ \ \ \ \ \ \ \ \ \ \ \ \ \ \ m:numExp]
			\ \ target t':arr <- join[a:arr,b:arr]
			\ \ antecedent ([n:numExp],[a:arr]) :: rep,
			\ \ \ \ \ \ \ \ \ \ \ \ \ ([m:numExp],[b:arr]) :: rep,
			\ \ \ \ \ \ \ \ \ \ \ \ \ ([],[a:arr,b:arr]) :: disj
			\ \ consequent ([t:numExp],[t':arr]) :: rep
	\end{alltt}}
\end{wrapfigure}
A transfer schema is declared in {\oruga} by specifying source and target patterns, and the antecedent and consequent constraints, as demonstrated here (right).

%
%

\vspace{-0.18cm}
\section{Structure Transfer}
\vspace{-0.12cm}
Structure transfer is a method for producing transformations of a given graph in a source construction space into some target construction space. The goal of structure transfer is to satisfy some constraint involving a given token and some sought-after token. For instance, if we start with token $1+2+3$ and we wish to find a dot arrangement which \textit{represents} it, structure transfer will use transfer schemas to try to build such dot arrangement while simultaneously proving that the desired constraint must hold.
A general version of this method is presented in \cite{raggi2022transfer}.

\begin{wrapfigure}[10]{r}{4.4cm}
	\vspace{-0.3cm}
	\begin{tikzpicture}[construction]
	\node[termrep] (v_{2}arr) at (0.0,0.1) {$\oT$};
	\node[constructor = {$\mathtt{join}$}] (join_v_{2}arr) at (-0.8,-1.0) {};
	\path[->] (join_v_{2}arr) edge (v_{2}arr);
	\node[termrep] (v_{4}arr) at (-1.3,-2.05) {$\oTsmall$};
	\path[->] (v_{4}arr) edge node[index label] {1} (join_v_{2}arr);
	\node[constructor = {$\mathtt{join}$}] (join_v_{4}arr) at (-1.3,-3.0) {};
	\path[->] (join_v_{4}arr) edge (v_{4}arr);
	\node[termrep,minimum width = 11.5pt,minimum height = 11.5pt] (v_{5}1arr) at (-1.8,-4.0) {$\crc$};
	\path[->] (v_{5}1arr) edge node[index label] {1} (join_v_{4}arr);
	\node[termrep] (v_{6}2arr) at (-0.7,-4.0) {$\oTwo$};
	\path[->] (v_{6}2arr) edge node[index label] {2} (join_v_{4}arr);
	\node[termrep] (v_{1}3arr) at (-0.15,-2.0) {$\oThree$};
	\path[->] (v_{1}3arr) edge node[index label] {2} (join_v_{2}arr);
	\node[constructorSW = {$\mathtt{rotate}$}] (rotateX_v_{2}arr) at (1.0,-0.75) {};
	\path[->] (rotateX_v_{2}arr) edge (v_{2}arr);
	\node[termrep] (v_{3}arr) at (1.2,-1.9) {$\oTrot$};
	\path[->] (v_{3}arr) edge node[index label] {1} (rotateX_v_{2}arr);
	\node[constructorW = {$\mathtt{remove}$}] (remove_v_{3}arr) at (1.2,-2.9) {};
	\path[->] (remove_v_{3}arr) edge (v_{3}arr);
	\node[termrep] (v_{0}arr) at (0.8,-4.0) {$\oR$};
	\path[->] (v_{0}arr) edge node[index label] {1} (remove_v_{3}arr);
	\path[->,in=15,out=0, looseness=1.1] (v_{2}arr) edge node[index label] {2} (remove_v_{3}arr);
	\end{tikzpicture}
\end{wrapfigure}
We have had success with various tests of structure transfer. For example, it is possible to define a transfer schema that roughly specifies that a dot arrangement represents an equation if the same arrangement represents each side of the equation. Thus, given $1+2+3 = 3(3+1)/2$, structure transfer will try to find one arrangement that can be constructed in two ways, one corresponding to $1+2+3$ and the other to $3(3+1)/2$. One result is a pair of constructions of arrangement \scalebox{0.75}{\oT}, as shown here (right). The generalisation of this result is a graphical proof of Gauss' sum.

\vspace{-0.18cm}
\section{Past and future work}
\vspace{-0.12cm}
{\oruga}'s purpose is to facilitate encoding diverse representations within a uniform framework so that we can perform transformations between them. So far, we have implemented a restricted version of the methods presented in~\cite{raggi2022transfer}, wherein transfer schemas are used as inference rules applied backwards (from the goal). Our approach is domain-independent.
To date we have used {\oruga} to transform across multiple construction spaces (e.g., arithmetic, Euler diagrams, set algebra, propositional logic and geometry). We discuss its generality in~\cite{raggi2022transfer}, and in particular its relation to similar but more specific formal methods~\cite{huffman2013lifting,reynolds1983types,coen2004semi}, as well as its relation to the application and discovery of analogies~\cite{gentner1983structure}. In future work we aim to explore more in depth the potential of structure transfer for analogy, and other applications of RST in cognitive science.
We are currently developing a graphical interface to improve its usability, especially for inputting constructions. 

\begin{acknowledgments}
	Supported by EPSRC grants EP/R030650/1, EP/T019603/1, EP/R030642/1, and EP/T019034/1.
\end{acknowledgments}
\bibliography{library}
\end{document}